\documentclass[letterpaper, 10 pt, conference]{ieeeconf}
    
    \IEEEoverridecommandlockouts
    \makeatletter
    \let\NAT@parse\undefined
    \makeatother
    \usepackage[numbers,sort&compress]{natbib}
    
    \usepackage[utf8]{inputenc}
    \usepackage[T1]{fontenc}
    \usepackage{courier}
    \usepackage[hidelinks,hypertexnames=false]{hyperref}
    \usepackage{url}
    \usepackage{booktabs}
    \usepackage{amsmath}
    \usepackage{amsfonts}
    \usepackage{graphicx}
    \usepackage{float}
    \usepackage{microtype}
    \usepackage{xcolor}
    \usepackage{placeins}
    \usepackage{adjustbox}
    \usepackage{multirow}

    \newcommand{\method}{SafeVantage}
    
    \newcommand{\support}{\mathrm{support}}

    \newcommand{\taskitem}[2]{%
       \item \texttt{#1}: #2%
     }
    
    \title{\LARGE \bf
    \method: Vantage-Aware Memory for Reliable Embodied Decisions
    }
    
    \author{Sean Hardesty Lewis$^{1}$, Zuyi Guo$^{2}$, Benwang Chen$^{2}$, Zirui Li$^{3}$, Hongyi Lin$^{4}$, and Heye Huang$^{2,\dagger}$%
    \thanks{$^{\dagger}$Corresponding author: {\tt\small heye.huang@kaist.ac.kr}.}%
    \thanks{This work was conducted as part of the MIT-UF-NEU Joint Summer Research Camp 2026.}%
    \\[1.5ex]
    $^{1}$Massachusetts Institute of Technology \\
    $^{2}$Korea Advanced Institute of Science and Technology \\ $^{3}$Nanyang Technological University \\ $^{4}$Tsinghua University
    }
    
    \usepackage{graphicx}
    \usepackage{etoolbox}
    \usepackage{caption}

    \newif\ifSVteaserregistered
    \SVteaserregisteredfalse    
    
    \makeatletter
    \apptocmd{\@maketitle}{%
        \par
        \vspace{1.0em}
    
        \ifSVteaserregistered
        \else
            \refstepcounter{figure}
            \label{fig:framework}
            \global\SVteaserregisteredtrue
        \fi
    
        \begin{center}
            \includegraphics[
                width=0.96\textwidth
            ]{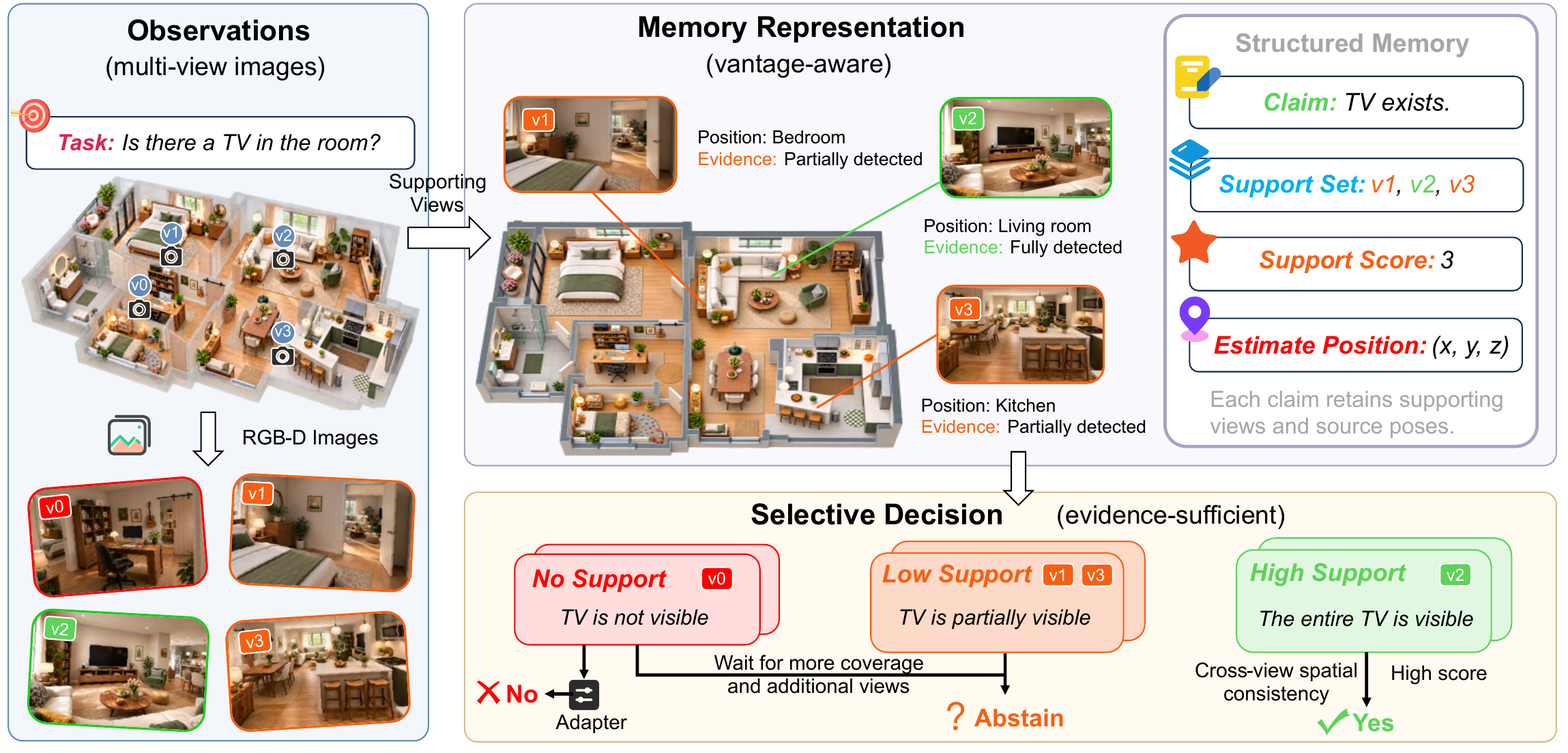}
    
            \vspace{0.3em}
    
            {\normalfont\normalsize
            \captionof*{figure}{%
            Fig.~\thefigure.\textbf{
            Overview of \method{}}, a vantage-aware memory for reliable embodied decisions. It retains the viewpoints that support each claim and uses the evidence to select the next view and make the final \textsc{Yes}/\textsc{No}/\textsc{ABSTAIN} decisions.
            }}
        \end{center}
    
        \vspace{0.8em}
    }{%
    }{%
        \PackageWarning{teaser}{Failed to modify \string\@maketitle}
    }
    \makeatother

\begin{document}
    
    \maketitle
    \thispagestyle{empty}
    \pagestyle{empty}

\begin{abstract}
Reliable embodied decisions under partial observability require informative observations and sufficient supporting evidence. However, semantic scores alone do not reveal which viewpoints justify a claim or where additional evidence should be acquired. We introduce \method{}, a vantage-aware semantic memory and active acquisition framework that retains each claim's supporting views, camera poses, and estimated target location, keeping positive support distinct from search coverage. A learned candidate-observability model uses claim-grounded geometry to predict target visibility at reachable viewpoints. These predictions guide view selection through expected reduction in terminal decision loss, accounting for travel cost and geometrically distinct corroboration. A calibrated head then combines support, spatial consistency, and coverage to produce \textsc{Yes}, \textsc{No}, or \textsc{Abstain} decisions. We evaluate \method{} on a category-presence benchmark spanning 232 unseen ProcTHOR houses and 7,424 paired episodes per method and action budget. Compared with validation-selected equal-budget baselines, \method{} achieves macro-F1 gains of 24.7\% and 12.0\% at eight and twelve actions, respectively, with lower risk and higher answer rates at both budgets and 31.7\% less travel at eight actions. Equal-input HM3D experiments show lower selective risk under fixed observations, while controlled ScanNet interventions show that restoring supporting views improves downstream VLM answers. Ablations further support the contribution of candidate observability to decision quality and acquisition efficiency. Results demonstrate the value of claim-level viewpoint evidence for connecting semantic memory, active acquisition, and reliable decision-making. Code is available at  \url{https://safevantage.github.io}
\end{abstract}
    

    \section{Introduction}
    
    Embodied agents increasingly rely on spatial memory to reason about
    objects and answer questions in partially observed environments.
    OpenScene, VLMaps, and ConceptGraphs connect language with persistent
    2D or 3D scene representations~\citep{peng2023openscene,
    huang2023vlmaps,gu2024conceptgraphs}, while recent EQA systems combine
    memory, exploration, and foundation models for embodied
    reasoning~\citep{das2018embodied,majumdar2024openeqa,exploreeqa2024}.
    However, reliable decision-making requires more than retrieving a likely
    semantic match. The memory must also indicate whether the available observations
    actually provide sufficient evidence for that claim.

    Existing approaches organize multi-view observations into maps, graphs, or selected snapshots~\citep{huang2023vlmaps,gu2024conceptgraphs,yang2024threedmem}. These representations support retrieval, but a semantic score alone does not reveal which viewpoints support a claim, whether they are spatially distinct, or how clearly the target was observed. These differences matter under partial observability: a single weak detection provides different evidence from corroborated observations, while missing detections may reflect incomplete search coverage rather than true absence. Embodied systems have used confidence to stop exploration or request help~\citep{geifman2019selectivenet,cheng2025efficienteqa}. These limitations motivate a claim-level evidence representation that preserves supporting observations, uses their provenance to guide subsequent acquisition, and distinguishes insufficient evidence from evidence of absence through selective abstention.

    In this work, we propose \method{}, a vantage-aware semantic memory that associates each claim with its supporting views, estimated target location, and unresolved evidence. The same evidence state drives the acquisition-to-decision loop: the policy explores for relevant evidence, uses claim-grounded target geometry to predict the observability of candidate views, and uses supporting-view provenance to seek geometrically distinct corroboration. The accumulated evidence then yields a selective \textsc{Yes}, \textsc{No}, or \textsc{Abstain} decision, distinguishing insufficient observation from evidence of absence. Fig.~\ref{fig:framework} illustrates the overview of \method{}.
    We evaluate \method{} on an active-view benchmark constructed from the official ProcTHOR-10K splits~\citep{deitke2022procthor}. Across 232 unseen houses and 7,424 paired episodes per method and action budget, \method{} improves macro-F1 and reduces decision risk against the strongest equal-budget baselines at both action budgets, while also reducing travel. We further isolate the role of vantage-aware evidence through equal-input HM3D comparisons, controlled ScanNet view-removal interventions, and component ablations.
    
    Our main contributions are summarized as follows:
    \begin{enumerate}
        \item We introduce \method{}, a vantage-aware semantic memory that
        preserves claim-level viewpoint provenance, target localization, and
        unresolved evidence instead of collapsing multi-view observations into
        a single semantic score.
    

        \item We develop a claim-conditioned active acquisition framework in which
        retained target geometry predicts candidate-view observability, while
        supporting-view provenance guides geometrically distinct corroboration
        before selective \textsc{Yes}/\textsc{No}/\textsc{ABSTAIN} decisions.
    
        \item We establish a ProcTHOR active-view evaluation with 5
        equal-budget acquisition baselines on which \method{} attains the best risk, macro-F1, and answer rate at both action budgets, complemented by HM3D equal-input evaluation, ScanNet interventions, and component ablations.
    \end{enumerate}

\begin{figure*}[t]
\centering
\includegraphics[width=0.96\textwidth]{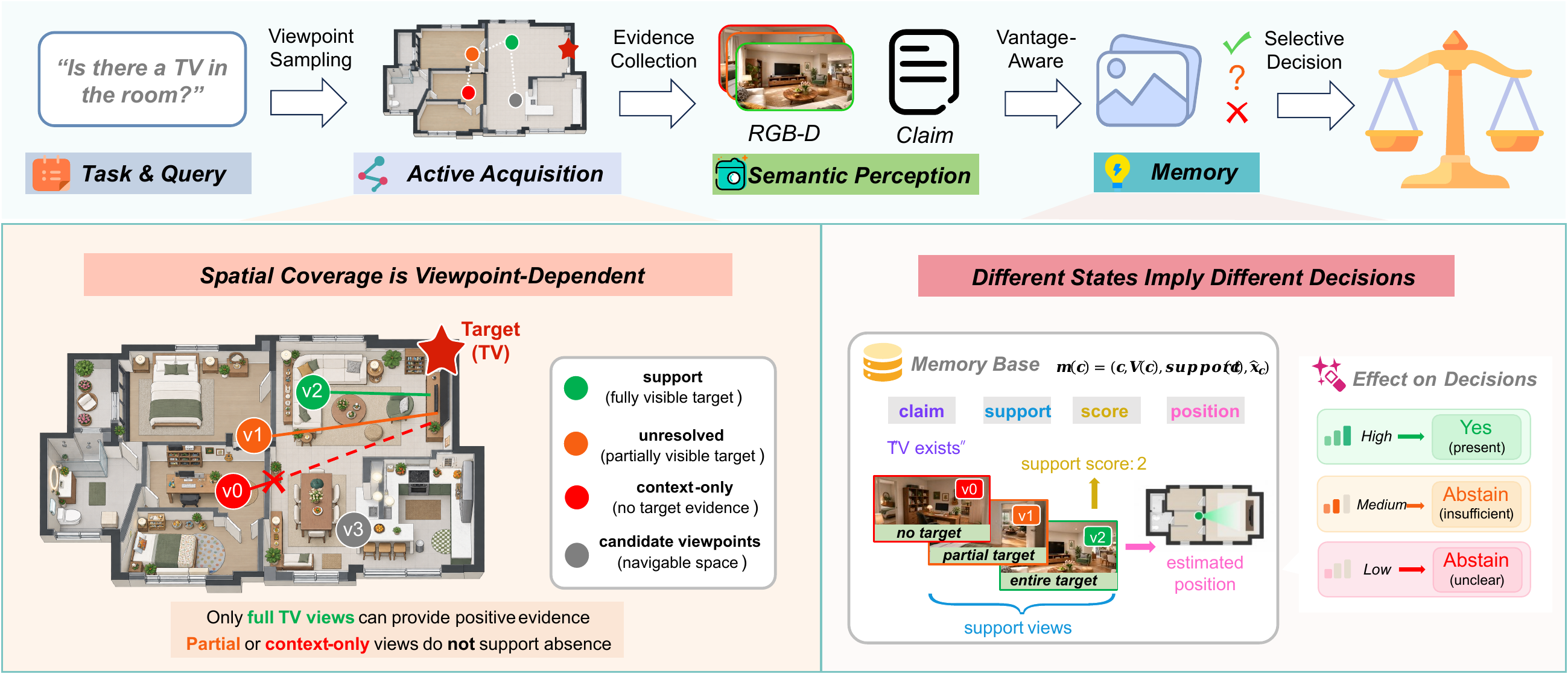}
\caption{\textbf{Viewpoint support and search coverage provide complementary evidence.} Navigable viewpoints differ in the target evidence they reveal. Supporting observations provide positive evidence, whereas partial or context-only observations do not justify absence. Retained views and target geometry inform subsequent acquisition and selective decisions.}
\label{fig:coverage}
\end{figure*}

\section{Related Work}

\subsection{Open-vocabulary Spatial Memory}

Open-vocabulary spatial representations connect language with scene geometry for semantic retrieval and navigation. VLMaps and OpenScene associate vision--language features with 2D maps and dense 3D geometry~\citep{huang2023vlmaps,peng2023openscene}, while LERF embeds language into radiance fields~\citep{kerr2023lerf}. ConceptGraphs and OpenVoxel organize observations into object-centric graphs and captioned voxel groups~\citep{gu2024conceptgraphs,huang2026openvoxel}. HOV-SG introduces floor--room--object hierarchies for navigation~\citep{werby2024hierarchical}, and FARM combines relational memory with viewpoint evidence~\citep{he2026farm}. These methods support semantic localization and retrieval, with some also retaining source observations. Rather than proposing another mapping backbone, \method{} uses claim-level viewpoint provenance for both active acquisition and selective decisions. ConceptGraphs and VLMaps serve as map-centric alternatives in our equal-input evaluation.

\subsection{Embodied Question Answering and Exploration}

EQA requires agents to collect and integrate observations under partial observability~\citep{das2018embodied}. OpenEQA benchmarks episodic and active answering, Explore-EQA calibrates exploration stopping, and EfficientEQA studies efficient open-vocabulary answering~\citep{majumdar2024openeqa,exploreeqa2024,cheng2025efficienteqa}. MemoryEQA, GraphEQA, and 3D-Mem organize observation histories, semantic scene graphs, and informative snapshots for retrieval and reasoning~\citep{memoryeqa2025,saxena2025grapheqa,yang2024threedmem}. EXPRESS-Bench evaluates exploration-grounded answers, while Mind Palace studies long-term active EQA with structured memory and value-of-information stopping~\citep{jiang2025express,ginting2025mindpalace}. DAAAM and UQ-DAAAM extend memory to dynamic scenes and uncertainty-aware refinement~\citep{gorlo2025daaam,zhang2026uqdaaam}. These approaches improve memory construction, retrieval, and exploration for answering questions. Our work focuses on category-presence decisions and uses each claim's supporting observations to guide further acquisition and assess whether a decision is sufficiently supported.

\subsection{Active Perception and View Selection}

Active perception selects observations to obtain task-relevant information. For semantic navigation, SemExp learns goal-oriented exploration from episodic semantic maps~\citep{chaplot2020object}, while PONI predicts potential functions over semantic maps to guide object search~\citep{ramakrishnan2022poni}. VLFM uses vision--language value maps for frontier selection~\citep{yokoyama2024vlfm}, and SG-Nav combines online 3D scene graphs and LLM reasoning with re-perception to address unreliable detections~\citep{yin2024sg}. These methods use semantic information to direct exploration and verify targets.
\method{} conditions acquisition on a claim-grounded evidence state: retained target geometry predicts candidate observability, and candidate views are selected according to expected reduction in terminal decision loss.
Our equal-budget baselines isolate coverage-driven, room-prior, and detector-confidence-guided acquisition rather than reproduce these complete navigation systems.

\subsection{Selective Prediction and Abstention}

Selective prediction introduces a reject option to trade coverage against prediction risk~\citep{geifman2019selectivenet, huang2026knowledge}. In robotics, KnowNo uses conformal prediction to determine when a planner should request assistance under ambiguity~\citep{ren2023knowno}. Explore-EQA calibrates exploration stopping~\citep{exploreeqa2024}, while AbstainEQA evaluates abstention when perceptual evidence is unavailable or underspecified~\citep{wu2026abstaineqa}. UQ-DAAAM further incorporates cross-view semantic uncertainty into memory refinement~\citep{zhang2026uqdaaam}. \method{} shares the abstention objective but focuses on the evidence supplied to the decision rule: supporting viewpoints and their spatial consistency remain explicit, while search coverage is represented separately from positive support. Controlled view-removal and restoration experiments further test how access to supporting evidence affects downstream VLM answers.

\section{Method}
\label{sec:method}

We consider category-presence queries under partial observability. Given a claim $c$, \method{} acquires observations within a fixed action horizon $B$ and a geodesic travel budget, then returns \textsc{Yes}, \textsc{No}, or \textsc{Abstain}. The framework comprises semantic memory, active acquisition, and selective decisions.

\subsection{Vantage-aware Semantic Memory}
\label{sec:vantage_memory}

At viewpoint $v_i$, the agent obtains an observation $o_i=(I_i,D_i,T_i,\tau_i)$ containing RGB, optional depth, camera pose, and timestamp. Let $H_t$ denote the observation history and $\delta_i(c)\in\{0,1\}$ indicate whether the perception adapter reports claim $c$. The supporting viewpoints are $V_t(c)=\{v_i \mid o_i\in H_t,\ \delta_i(c)=1\}$ with support count $\support_t(c)=|V_t(c)|$. Each supporting viewpoint retains its source observation and camera pose. When depth is available, back-projection provides per-view target estimates $\hat{\mathbf{x}}_c^{(i)}$ and an approximate aggregate location $\hat{\mathbf{x}}_{c,t}$. The memory record is $m_t(c)=(c,V_t(c),\support_t(c),\hat{\mathbf{x}}_{c,t})$, where the target location is optional. Because the consistency gate (Eq.~\ref{eq:pair_gate}) accepts only pairs of per-view estimates within radius $r$, detections of instances separated by more than $r$ cannot corroborate one another. The estimate $\hat{\mathbf{x}}_{c,t}$ summarizes the supporting detections by their scores and is used to score candidate views; it does not assume a unique object instance. Support is adapter-specific: ProcTHOR uses YOLO-World scores and boxes~\citep{cheng2024yoloworld}, while HM3D uses Qwen2-VL-7B per-view object lists~\citep{wang2024qwen2vl}. Simulator semantics are not queried by the test-time policy; target visibility is additionally used as a training target for the candidate-observability model. Support counts record detections; their spatial diversity and geometric agreement are assessed separately.

Target visibility, property or relation sufficiency, search coverage, and answer correctness are distinct. For category presence, supporting detections provide positive evidence, whereas missing support may reflect incomplete observation rather than absence (Fig.~\ref{fig:coverage}). Coverage therefore enters the decision features separately from support. Retained source observations also enable the controlled ScanNet interventions described in the experiments.





\subsection{Claim-conditioned Active Acquisition}
\label{sec:active_acquisition}

Let $\mathcal{A}_t$ contain unvisited viewpoints reachable within the
remaining geodesic budget, and let $d_t(a)$ denote the incremental
geodesic travel distance to candidate $a$ in meters.

For each candidate $a$, \method{} predicts whether the target is likely to
be observable from that viewpoint. We form a geometric feature vector
$\psi_t(a,c)$ from the current claim state and candidate geometry (distance
to the estimated target, heading alignment, and a geodesic baseline
feature), standardize it with statistics frozen from the training scenes,
and score it with a logistic outcome model
$\hat{o}_t(a,c)=\sigma(\boldsymbol{\theta}^{\top}\tilde{\psi}_t(a,c)+b_o)$,
fit only on training scenes using target observability as its label. At test
time the selector sees only the acquired observations, the resulting claim
state, and candidate geometry; oracle visibility, simulator instance
identity, and test labels are never queried.

Candidate views are selected by their expected reduction in terminal
decision loss. Let $\hat{p}_t$ be the calibrated belief that $c$ holds given
the current top supporting score. A candidate yields a positive reading with
probability $p_t^{+}=\hat{p}_t\hat{o}_t(a,c)+(1-\hat{p}_t)\phi$, where $\phi$
is a fixed detector false-positive rate; each outcome induces a Bayes
posterior $q_t^{+}=\hat{p}_t\hat{o}_t/p_t^{+}$,
$q_t^{-}=\hat{p}_t(1-\hat{o}_t)/(1-p_t^{+})$ and a hypothetical evidence
state $H_t^{+}$ or $H_t^{-}$ formed with the online update used at execution.
Under decision $y\in\{\textsc{Yes},\textsc{No},\textsc{Abstain}\}$ and belief
$q$, the terminal loss
$\ell(y,q)=5(1-q)\mathbf{1}[y{=}\textsc{Yes}]+2q\,\mathbf{1}[y{=}\textsc{No}]+0.5\,\mathbf{1}[y{=}\textsc{Abstain}]$
is the per-decision form of the risk in Eq.~\ref{eq:risk_cost}. Applying the
selective rule $\pi_{\mathrm{active}}$ (Eq.~\ref{eq:active_decision}) to each
state, the next viewpoint maximizes the travel-discounted expected loss
reduction:
\begin{equation}
a_t^*=\arg\max_{a\in\mathcal{A}_t}
\frac{L_t(c)-\bar{L}_t(a;c)-\lambda_d\,d_t(a)}{0.5+d_t(a)},
\quad \lambda_d=0.001,
\label{eq:view_selection}
\end{equation}
where $L_t(c)=\ell(\pi_{\mathrm{active}}(H_t),\hat{p}_t)$ is the
no-acquisition loss and
$\bar{L}_t(a;c)=p_t^{+}\,\ell(\pi_{\mathrm{active}}(H_t^{+}),q_t^{+})+(1-p_t^{+})\,\ell(\pi_{\mathrm{active}}(H_t^{-}),q_t^{-})$
is the expected loss after acquiring $a$. Each acquired observation updates
the claim state before the next candidate is scored; action horizons and
geodesic travel limits are fixed by the evaluation protocol.

\noindent\textit{Exploration before grounding.}
Before a grounded supporting detection provides an estimate of the target location, the target-dependent features in $\psi_t(a,c)$ are unavailable. The policy therefore explores using spatial novelty, category--room evidence, and travel cost to score reachable candidate views. This exploration branch remains active throughout episodes in which the category is absent. Once a supporting detection yields a target estimate $\hat{\mathbf{x}}_{c,t}$, the policy switches to the candidate-observability selector in Eq.~\ref{eq:view_selection}. Observations with no detection increase search coverage, while grounded positive detections update the supporting-view set and target estimate. Neither branch uses oracle visibility or simulator instance identity at test time.

\subsection{Evidence-based Selective Decisions}
\label{sec:selective_decisions}

\noindent\textit{Active decision head.} At the fixed horizon, the feature vector $\phi(H_B,c)$ summarizes the top three detection scores from distinct positions, support counts at five thresholds, observed-position coverage, room and observation counts, grounded-evidence fraction, minimum inter-view target distance, maximum context evidence, and category identity. An $\ell_2$-regularized logistic model estimates category-presence probability:
\begin{equation}
p_B(c)=\sigma\!\left(\mathbf{w}^{\top}\phi(H_B,c)+b\right),
\label{eq:presence_probability}
\end{equation}
where $\sigma(z)=(1+\exp(-z))^{-1}$. The parameters $\mathbf{w}$ and $b$ are fitted separately for each policy using training-only observations and five scene-disjoint folds.

Positive commitments additionally require geometrically consistent support. Let $\mathcal{P}_B(c)$ contain observation pairs meeting the support threshold, originating from distinct positions, and providing valid target estimates. The consistency gate is
\begin{equation}
g_B(c)=\mathbf{1}\!\left[
\exists(i,j)\in\mathcal{P}_B(c):
\left\|\hat{\mathbf{x}}_c^{(i)}-\hat{\mathbf{x}}_c^{(j)}\right\|_2\leq r
\right],
\label{eq:pair_gate}
\end{equation}
where $r$ is the fixed agreement radius. The decision rule is
\begin{equation}
\pi_{\mathrm{active}}(H_B,c)=
\begin{cases}
\textsc{Yes}, & p_B(c)\geq\alpha\ \land\ g_B(c)=1,\\
\textsc{No}, & p_B(c)\leq\beta,\\
\textsc{Abstain}, & \text{otherwise},
\end{cases}
\label{eq:active_decision}
\end{equation}
with $0\leq\beta<\alpha\leq1$. Thresholds for \method{} are selected by nested validation and frozen before testing. Coverage contributes through the feature vector rather than guaranteeing absence.


\noindent\textit{Static memory adapter.} For equal-input HM3D comparisons, each method's scalar query score is mapped to \textsc{Yes}/\textsc{No}/\textsc{Abstain} by the same two-threshold adapter, calibrated on the calibration scenes and fixed for evaluation; ties favor higher answer coverage, then lower false-positive rate.

\noindent\textit{Decision cost.} We evaluate incorrect commitments, abstention, and resource use through
\begin{equation}
\begin{aligned}
R={}&5\,\mathbf{1}[\mathrm{FP}]+2\,\mathbf{1}[\mathrm{FN}]\\
&+0.5\,\mathbf{1}[\mathrm{reject}]+0.001\,C_{\mathrm{resource}}.
\end{aligned}
\label{eq:risk_cost}
\end{equation}
$\mathrm{FP}$ and $\mathrm{FN}$ denote incorrect positive and negative commitments; $\mathrm{reject}$ denotes \textsc{Abstain}. False positives receive the largest penalty. $C_{\mathrm{resource}}$ is travel in meters for active acquisition and view count for static evaluation. This evaluation cost is distinct from the logistic training objective.

    \section{Experiments}
    \subsection{Experimental Setup}
    \textbf{ProcTHOR benchmark.}
    Our primary evaluation uses official ProcTHOR-10K train, validation, and test identities~\citep{deitke2022procthor} and derives a 24-category vocabulary from training metadata.  Every scene provides 40 navigable positions with 4 camera yaws, yielding a common 160-view action graph with geodesic edge costs. For each eligible scene, the evaluator constructs a balanced query set by sampling 4 present and 4 absent categories and assigning 4 deterministic starts to each category, producing 32 paired episodes per scene. Of 256 held-out identities, 255 materialized successfully; 232 satisfied the balanced-query construction, yielding 7,424 episodes per method and action budget. All policies receive RGB-D observations, poses, the training-derived vocabulary, and YOLO-World scores~\citep{cheng2024yoloworld}. Oracle visibility and simulator instance identity are not queried by the test-time selector; visibility is additionally used as a training-only target for the candidate-observability model and for evaluation.

    \textbf{Baselines.} 
    To comprehensively evaluate \method{}, we select 5 baselines:

    \begin{itemize}
       \setlength{\itemsep}{0pt}
       \setlength{\parskip}{0pt}
       \setlength{\parsep}{0pt}

       \taskitem{Random}
       {Select a deterministic hash-random candidate.}

       \taskitem{Nearest Frontier}
        {Minimize incremental geodesic distance.}

       \taskitem{Coverage Gain}
       {Maximize newly covered positions.}

       \taskitem{Category–Room Prior}
       {Combine coverage with a training-derived room model.}

       \taskitem{Detector-Confidence Gain}
       {Seek a distinct nearby view around the strongest detection.}
    \end{itemize}

    Different policies induce different distributions over observation histories, so 5 scene-disjoint folds are used to fit a selective head for each policy. Nested validation then selects category–room prior at 8 actions and detector-confidence gain at 12 actions as the primary comparison policies. \method{} uses the claim-grounded memory state to predict candidate observability and selects viewpoints according to expected reduction in terminal decision loss under fixed sensing and travel budgets.
    
  
    \textbf{Metrics.}
    All ProcTHOR comparisons are paired by scene, category, start, and budget. We report task risk, macro-F1 over positive and negative decisions, false-positive rate on absent queries, answer rate, accuracy conditional on answering, and mean geodesic travel. Macro-F1 is the mean of the \textsc{Yes} and \textsc{No} F1 scores; an \textsc{Abstain} is not counted as a positive or negative prediction and acts as a miss for the true class, so abstaining lowers recall and cannot inflate macro-F1. We estimate confidence intervals using 10,000 paired bootstrap draws that resample whole scenes, preserving the 32 correlated episodes within each house.
    

    
    \subsection{Held-out Acquisition-to-decision Performance}
    

    We compare \method{} with five acquisition baselines at both budgets using shared scenes, detectors, action graphs, and decision heads (Table~\ref{tab:locked-all-baselines}).
    
    \begin{table}[t]
    \centering
    \caption{
    Equal-budget acquisition-to-decision performance on 232 held-out ProcTHOR houses. Lower risk and travel are better; higher macro-F1 and answer rate are better. Bold marks the best value within each budget. Travel is measured in meters. $^{\dagger}$ represents the primary comparison policy.
    }
    \label{tab:locked-all-baselines}

    \renewcommand{\arraystretch}{1.10}
    \setlength{\tabcolsep}{3.2pt}

    \resizebox{0.98\columnwidth}{!}{%
    \begin{tabular}{@{}clcccc@{}}
    \toprule
    Budget
    & Method
    & Risk$\downarrow$
    & Macro-F1$\uparrow$
    & Answer$\uparrow$
    & Travel$\downarrow$ \\
    \midrule


    \multirow[c]{6}{*}{8}
    & \textbf{SafeVantage}
    & \textbf{0.384}
    & \textbf{0.604}
    & \textbf{0.581}
    & 4.94 \\

    & Random
    & 0.427
    & 0.434
    & 0.350
    & 19.12 \\
    
    & Nearest frontier
    & 0.398
    & 0.491
    & 0.394
    & \textbf{3.46} \\

    & Coverage gain
    & 0.422
    & 0.406
    & 0.305
    & 6.31 \\
    
    & Category-room prior $^{\dagger}$
    & 0.396
    & 0.484
    & 0.377
    & 7.23 \\

    & Detector-conf. gain
    & 0.407
    & 0.446
    & 0.333
    & 5.94 \\
    
    \addlinespace[2pt]
    \midrule
    

    \multirow[c]{6}{*}{12}
    & \textbf{SafeVantage}
    & \textbf{0.365}
    & \textbf{0.656}
    & \textbf{0.639}
    & 7.44 \\

    & Random
    & 0.420
    & 0.451
    & 0.361
    & 19.25 \\

    & Nearest frontier
    & 0.395
    & 0.464
    & 0.337
    & \textbf{5.28} \\
    
    & Coverage gain
    & 0.385
    & 0.534
    & 0.471
    & 8.97 \\
    
    & Category-room prior
    & 0.382
    & 0.490
    & 0.356
    & 8.16 \\

    & Detector-conf. gain $^{\dagger}$
    & 0.369
    & 0.585
    & 0.516
    & 8.42 \\
    
    \bottomrule
    \end{tabular}%
    }
    
    \end{table}
    At 8 actions, \method{} achieves the lowest risk and the highest macro-F1 and answer rate among all evaluated policies. Relative to the validation-selected category-room prior, macro-F1 increases from 0.4845 to 0.6044, an absolute improvement of 0.1199, and answer rate increases from 0.3772 to 0.5807. Mean travel decreases from 7.23 to 4.94~m, corresponding to a reduction of 31.7\%. Mean risk decreases from 0.3964 to 0.3842. The lower risk together with substantially higher answer rate shows that the improvement is not obtained by increasing the frequency of \textsc{ABSTAIN} decisions.
    At 12 actions, detector-confidence gain is the validation-selected comparator. \method{} increases macro-F1 from 0.5855 to 0.6559, an absolute improvement of 0.0704, and answer rate increases from 0.5163 to 0.6389. Mean risk decreases from 0.3685 to 0.3649, while mean travel decreases from 8.42 to 7.44~m.

\begin{figure}[t]
      \centering
      \includegraphics[width=\linewidth]{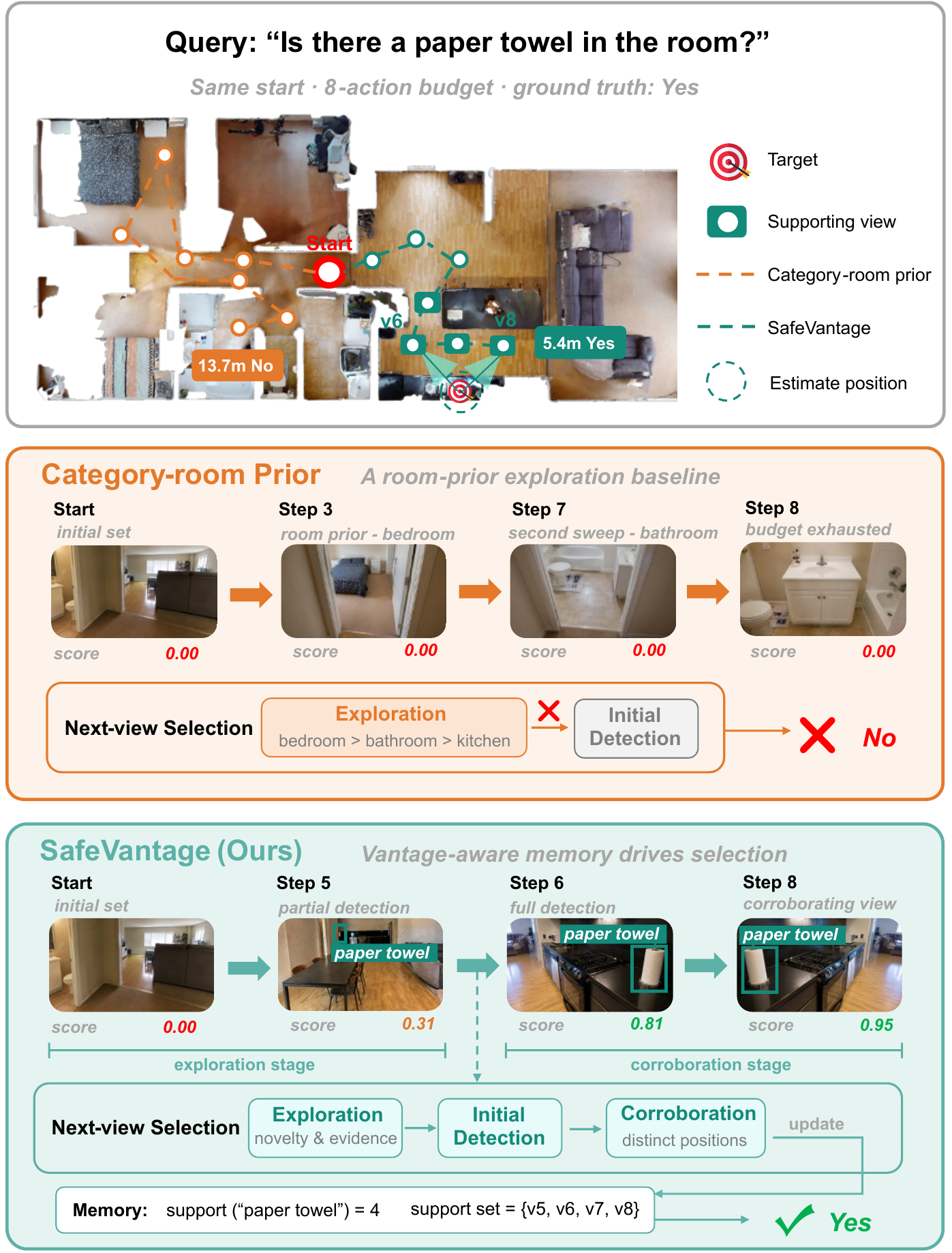}
      \caption{\textbf{One held-out episode for “Is there a paper towel in the room?”} Both policies start from the same start with an 8-action budget and a positive ground-truth label. Category–room prior does not observe the target, answers \textsc{NO}, and travels 13.7 m; \method{} acquires 4 supporting views, answers \textsc{YES}, and travels 5.4 m.}
      \label{fig:locked-traces}
    \end{figure}

    The comparison with all five baselines shows the same overall pattern. \method{} ranks first in risk, macro-F1, and answer rate at both action budgets, and it travels less than four of the five baselines. Nearest frontier is the only method with lower travel, but  the reduction is accompanied by substantially weaker decisions. These results highlight that minimizing travel alone does not necessarily yield evidence that supports reliable semantic decisions. In contrast, effective acquisition must balance movement efficiency with the informativeness of the observations collected.
    Fig.~\ref{fig:locked-traces} illustrates one held-out episode in which \method{} converts an initial partial detection into corroborated evidence and answers correctly while travelling less than the comparison policy.
    
    \begin{table*}[t]
    \centering
        \caption{
            Equal-input HM3D experiments. $R_9$ repeats calibration across nine folds; $R_{@\mathrm{cov}}$ matches coverage. Lower risk and AURC are better.
        }
        \label{tab:main}
    
        \small
        \renewcommand{\arraystretch}{1.15}
        \begin{tabular*}{\textwidth}{@{\extracolsep{\fill}}lrrrrrrrr@{}}
            \toprule
            Method
            & Cov.
            & Acc.
            & FP
            & Recall
            & Risk
            & $R_9$
            & $R_{@\mathrm{cov}}$
            & AURC \\
            \midrule
    
            \textbf{SafeVantage}
            & 0.436
            & \textbf{0.962}
            & \textbf{0.059}
            & 0.584
            & \textbf{0.524}
            & \textbf{0.538}
            & \textbf{0.538}
            & \textbf{0.649} \\
    
            Qwen2 caption-RAG
            & 0.565
            & 0.881
            & 0.240
            & \textbf{0.692}
            & 0.715
            & 0.638
            & 0.685
            & 0.715 \\
    
            ConceptGraphs
            & 0.432
            & 0.920
            & 0.123
            & 0.552
            & 0.617
            & 0.594
            & 0.707
            & 0.726 \\
    
            VLMaps
            & 0.541
            & 0.916
            & 0.162
            & 0.689
            & 0.618
            & 0.656
            & 0.910
            & 0.850 \\
    
            \bottomrule
        \end{tabular*}
    \end{table*}

    \subsection{Secondary Evidence in HM3D and ScanNet}
    
    Having tested the complete policy under the ProcTHOR benchmark, we conduct two secondary studies that examine supporting-view evidence under fixed observations and controlled interventions. On HM3D~\citep{yadav2023hm3dsem}, we compare semantic memory methods using identical RGB-D/pose inputs and a shared selective-decision rule. On ScanNet~\citep{dai2017scannet}, we remove and restore geometrically verified target-visible views while preserving the remaining input conditions, measuring their effect on downstream VLM answers.

    \textbf{Equal-input evaluation on HM3D.}
    We compare \method{}, Qwen2 caption-RAG, ConceptGraphs, and VLMaps on 36 HM3D-Sem scenes and 1,436 category-presence queries. Every method receives the same 160 RGB-D observations and camera poses per scene, with calibration repeated over nine disjoint 4-scene subsets. At the minimum-risk operating point, \method{} attains risk 0.524, versus 0.715, 0.617, and 0.618 for caption-RAG, ConceptGraphs, and VLMaps, and also the lowest out-of-calibration risk, matched-coverage risk, and AURC (Table~\ref{tab:main}).

    
    \textbf{Supporting-view intervention on ScanNet.} To test whether supporting observations materially affect downstream answers, we conduct a controlled intervention on 73 ScanNet questions. Official instance geometry, camera poses, and depth determine whether the queried instance is visible, independently of the VLM reader. For each question, target-visible frames are replaced at their original ranks by target-absent frames while the question, reader, prompt configuration, input ranks, and history length remain fixed. Starting from this target-absent history, we restore the strongest supporting view, the strongest 4 views, or the complete original supporting-view set, thereby varying the available target evidence while preserving the input-history structure. As shown in Table~\ref{tab:causal}, Qwen2-VL token F1 increases from 0.2483 without target-visible views to 0.4410 with 4 restored views and 0.4540 with the complete set; Qwen2.5-VL improves from 0.1888 to 0.3215 under complete restoration. In a separate negative-decision diagnostic, we combine view-opportunity and source-group coverage with evidence scores, calibrating the threshold on 6 development scenes under a 5\% false-negative-decision constraint and evaluating on 12 held-out ScanNet scenes. The rule achieves 0.983 observed-absent recall and 0.935 negative-decision precision, versus 0.051 and 0.500 for a score-only rule, with a 0.033 erroneous-negative-decision rate over non-absent states. Together, these results show that supporting views improve downstream answers, while coverage helps distinguish observed absence from unresolved missing evidence.

    \begin{table}[t]
      \centering
      \caption{ScanNet supporting-view intervention. Intervals compare restored conditions with target-absent histories.}
      \label{tab:causal}
      \small
      \setlength{\tabcolsep}{2pt}
      \renewcommand{\arraystretch}{1.18}
      \begin{tabular}{llccc}
\toprule
reader & evidence set & F1 & $\Delta$ vs absent & scene CI \\
\midrule
Qwen2-VL & target absent & 0.2483 & -- & -- \\
Qwen2-VL & strongest 1 & 0.3565 & +0.1082 & [-0.0181, 0.2290] \\
Qwen2-VL & strongest 4 & 0.4410 & +0.1927 & [0.0691, 0.2977] \\
Qwen2-VL & complete & \textbf{0.4540} & \textbf{+0.2057} & \textbf{[0.0752, 0.3133]} \\
Qwen2.5-VL & target absent & 0.1888 & -- & -- \\
Qwen2.5-VL & complete & 0.3215 & +0.1327 & [0.0513, 0.2053] \\
\bottomrule
\end{tabular}

    \end{table}

    \begin{figure}[t]
      \centering
      \includegraphics[
        width=\linewidth,
        height=0.25\textheight,
        keepaspectratio
      ]{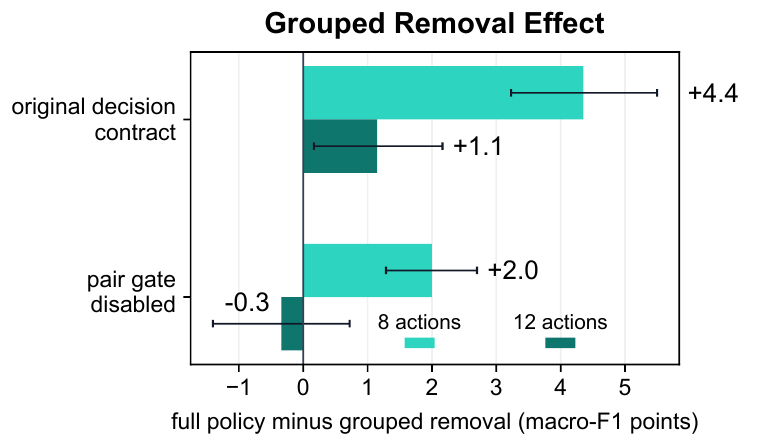}
      \caption{\textbf{Removing vantage-aware state lowers macro-F1.} Bars show full policy minus grouped removal on 96 validation houses; whiskers are paired 95\% intervals.}
      \label{fig:grouped-vantage}
    \end{figure}

\begin{table}[b]
  \centering
   \caption{Component ablations on 96 ProcTHOR validation houses. Loss is the macro-F1 drop from the full model (pp).}
  \label{tab:component-validation}

  \small
  \setlength{\tabcolsep}{2pt}
  \renewcommand{\arraystretch}{1.18}

  \begin{tabular*}{\columnwidth}{@{\extracolsep{\fill}}lcccc@{}}
  \toprule
  & \multicolumn{2}{c}{8 actions}
  & \multicolumn{2}{c}{12 actions} \\
  \cmidrule(lr){2-3}\cmidrule(l){4-5}

  Variant
  & F1$\uparrow$ & Loss (pp)$\downarrow$
  & F1$\uparrow$ & Loss (pp)$\downarrow$ \\
  \midrule

  \textbf{Full \method{}}
  & \textbf{0.5307} & --
  & \textbf{0.6149} & -- \\

  $-$ claim provenance
  & 0.4753 & \textbf{+5.54}
  & 0.5826 & \textbf{+3.23} \\

  $-$ target-ray alignment
  & 0.4749 & \textbf{+5.58}
  & 0.5819 & \textbf{+3.30} \\

  $-$ yaw diversity
  & 0.5311 & $-0.04$
  & 0.6162 & $-0.13$ \\

  $-$ base geometry
  & 0.5255 & \textbf{+0.52}
  & 0.6167 & $-0.18$ \\

  $-$ directed exploration
  & 0.5247 & \textbf{+0.60}
  & 0.4923 & \textbf{+12.26} \\

  \addlinespace[2pt]
  Confidence backbone only
  & 0.4847 & \textbf{+4.60}
  & 0.4835 & \textbf{+13.14} \\

  \bottomrule
\end{tabular*}
\end{table}
    
    \subsection{Ablations}

    We finally examine which acquisition components contribute to decision quality and how their effects depend on the final evidence gate. All variants are evaluated on 96 query-eligible ProcTHOR validation houses at 8- and 12-action budgets under a shared 20~m travel cap, yielding 3,072 paired episodes per variant and budget.


    \textbf{Grouped removal.} Jointly removing supporting-view identity, target-ray alignment, and yaw diversity (retaining room evidence, coverage, base geometry, and the travel objective) lowers macro-F1 by 4.35/1.15 points at 8/12 actions (95\% CIs $[3.23,5.50]$/$[0.16,2.16]$; Fig.~\ref{fig:grouped-vantage}). The effect persists with the pair gate disabled in both policies, separating acquisition from decision-time consistency.

    \textbf{Candidate observability.}
    Removing the candidate-observability model while holding the remaining
    controller fixed lowers macro-F1 by 7.52 and 8.03 points at 8 and
    12 actions, respectively. Mean travel simultaneously increases from
    4.94 to 6.30~m and from 7.44 to 8.95~m. So, candidate observability
    improves both decision quality and acquisition efficiency across both
    action horizons.
    
    \textbf{Component contributions.} Table~\ref{tab:component-validation} reports 5 single-component removals and a confidence-backbone-only variant. Removing supporting-view identity or target-ray alignment reduces macro-F1 by 5.54/3.23 and 5.58/3.30 points at 8/12 actions, respectively. Directed exploration has a strongly budget-dependent contribution: removing it costs 0.60 points at 8 actions but 12.26 points at 12 actions. Retaining only the confidence backbone loses 4.60/13.14 points. Other geometric terms show smaller or inconsistent marginal effects.
    

\subsection{Evaluation on a Disjoint Holdout}

To test generalization beyond the original 232-house evaluation, we held the method, thresholds, and evaluator fixed and tested on a disjoint cohort of 233 additional ProcTHOR houses that were never used for training, validation, or the original evaluation. On this cohort, \method{} improves macro-F1 by 7.3 and 4.3 points at 8 and 12 actions, respectively (95\% paired scene-bootstrap intervals $[6.4,8.1]$ and $[3.3,5.2]$). Risk decreases by 0.024 and 0.015, respectively, with higher answer rate and
lower travel at both horizons. These results confirm that the
acquisition advantage transfers to unseen houses.

\section{Limitations}

Our study focuses on category-presence queries, providing a controlled setting for examining the role of viewpoint evidence. Extending this formulation to attributes, relations, and temporal changes may require richer representations of evidence sufficiency. The acquisition policy uses a fixed perception stack and a candidate-observability model, with observations selected from a discrete viewpoint graph under fixed action budgets. Robustness to changes in perception and sensing conditions therefore remains an area for further evaluation. While the HM3D and ScanNet studies provide complementary evidence analyses, continuous navigation and real-robot evaluation are natural next steps toward assessing the framework under broader operating conditions.

\section{Conclusion}

In this work, we addressed the problem of acquiring sufficient visual evidence for reliable embodied decisions under partial observability. We introduced \method{}, a vantage-aware semantic memory that retains each claim's supporting views and target geometry while keeping positive support distinct from search coverage. This retained vantage evidence is what enables the rest of the system: it grounds the learned candidate-observability model, drives view selection by expected reduction in decision loss, and, through a geometric consistency gate, calibrated \textsc{Yes}/\textsc{No}/\textsc{Abstain} decisions. Experiments on held-out ProcTHOR houses show higher macro-F1 and answer rates with lower risk and travel than the validation-selected equal-budget baselines at both action horizons. Complementary HM3D comparisons demonstrate lower selective risk under identical observations, while ScanNet interventions show that restoring supporting views improves downstream VLM answers. Ablations further support the contributions of candidate observability and claim provenance. Together, these results support treating memory as an evidence state that guides both observation acquisition and semantic commitments, rather than solely as a representation for retrieval.

\section{Acknowledgments}

The authors used generative AI tools during manuscript preparation to assist with generating and revising selected text and figures, as well as code generation and debugging.

This work was supported by the National Research Foundation of Korea (NRF) grant funded by the Korean government (MSIT) under the project “Development of Risk-Enhanced Continual Learning for Embodied Intelligence in Long-Tail Environments” (Grant No. RS-2026-25595451) and by the Korea Institute of Science and Technology Information (KISTI) R\&D program through the joint research project “Development of the Next-Generation Integrated Wired/Wireless Communication Gateway (X-Gateway).”

    
    
    \bibliographystyle{IEEEtran}
    \bibliography{references}
    
    \clearpage

    \end{document}